\documentclass{bmvc2k}
\usepackage{hyperref}

\title{Inside Out Visual Place Recognition}

\addauthor{Sarah Ibrahimi \\ Nanne van Noord \\ Tim Alpherts \\ Marcel Worring}
{{s.ibrahimi, n.j.e.vannoord, t.o.l.alpherts, m.worring}@uva.nl}{1}

\addinstitution{
University of Amsterdam\\
Science Park 904\\
Amsterdam\\
The Netherlands}

\runninghead{Ibrahimi et al.}{Inside Out Visual Place Recognition}

\begin{document}

\maketitle

\begin{abstract}
Visual Place Recognition (VPR) is generally concerned with localizing outdoor images. However, localizing indoor scenes that contain part of an outdoor scene can be of large value for a wide range of applications. In this paper, we introduce \textit{Inside Out Visual Place Recognition (IOVPR)}, a task aiming to localize images based on outdoor scenes visible through windows. For this task we present the new large-scale dataset \textit{Amsterdam-XXXL}, with images taken in Amsterdam, that consists of 6.4 million panoramic street-view images and 1000 user-generated indoor queries. Additionally, we introduce a new training protocol \textit{Inside Out Data Augmentation} to adapt Visual Place Recognition methods for localizing indoor images, demonstrating the potential of Inside Out Visual Place Recognition. We empirically show the benefits of our proposed data augmentation scheme on a smaller scale, whilst demonstrating the difficulty of this large-scale dataset for existing methods. With this new task we aim to encourage development of methods for IOVPR. The dataset and code are available for research purposes at \url{https://github.com/saibr/IOVPR}. 
\end{abstract}

\section{Introduction}

\begin{figure}
\begin{tabular}{cc}
\centering
\includegraphics[width=7.0cm]{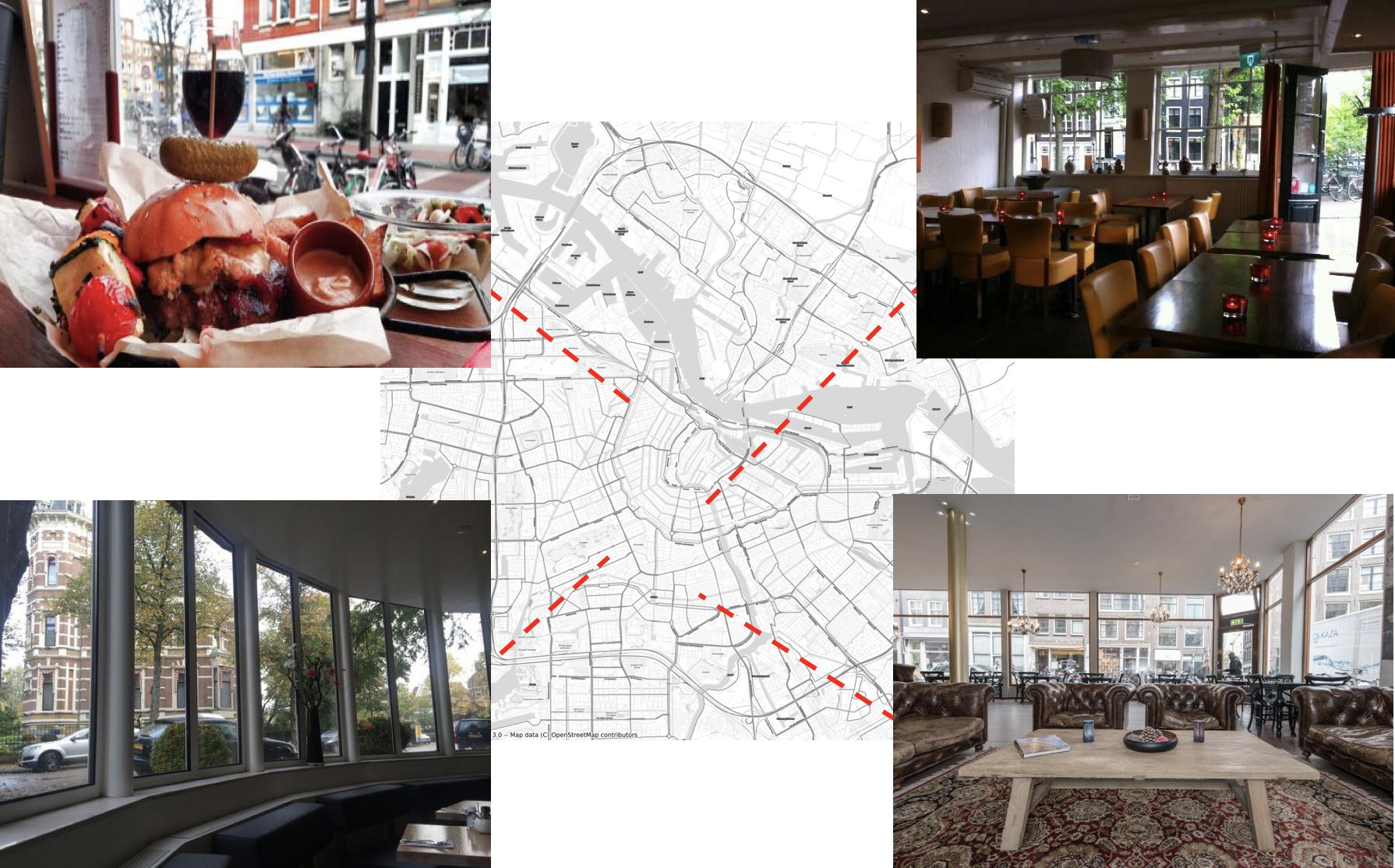}&
\includegraphics[width=4.5cm]{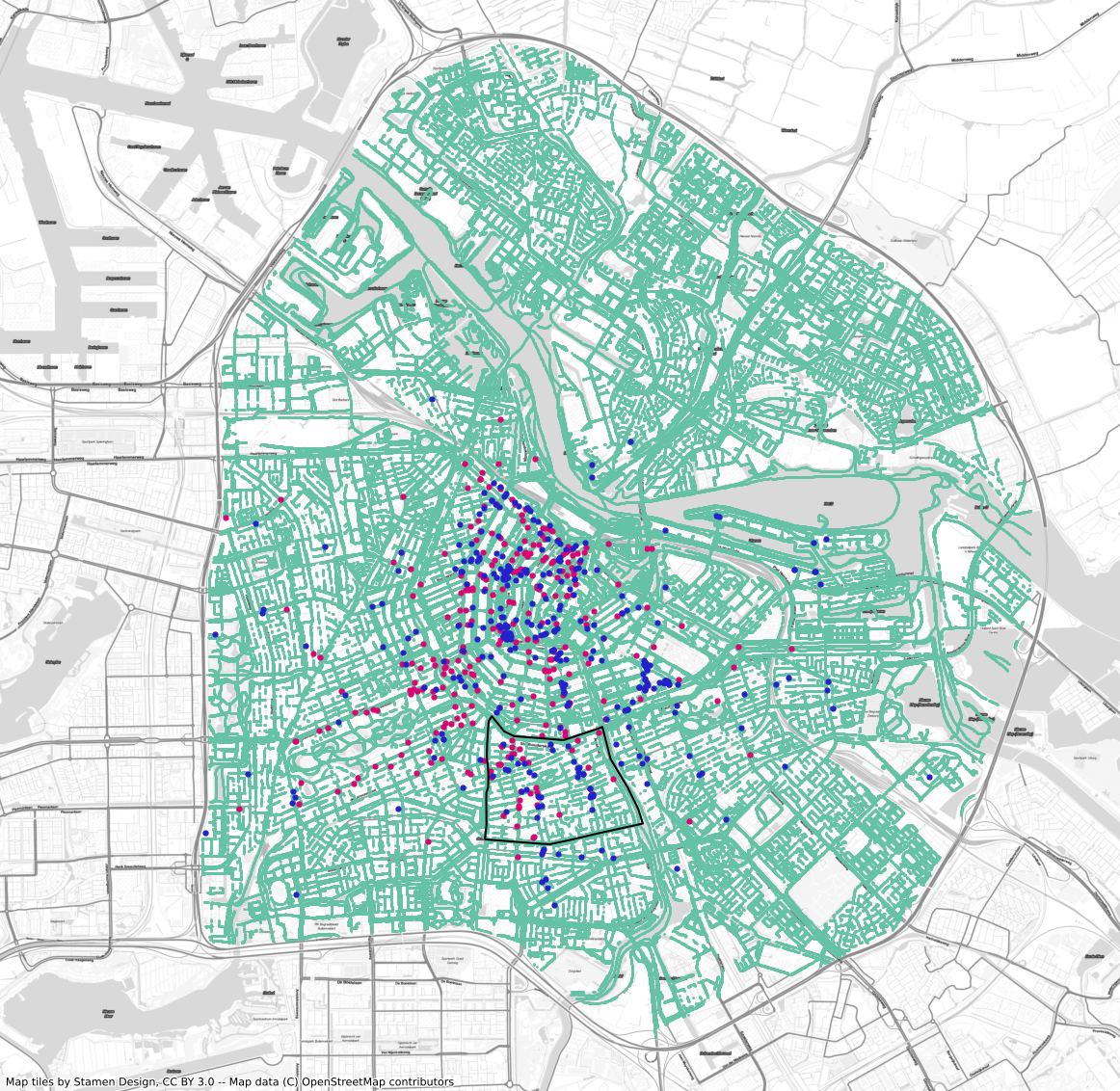}\\
(a)&(b)
\end{tabular}
\caption{(a) Examples of images taken in Amsterdam that are localized in the Inside Out Visual Place Recognition task. (b) Visualization of the Amsterdam-XXXL dataset. The green points indicate the Outdoor-Ams locations. The blue and pink points are the locations of the query images from Indoor-Ams validation set and test set respectively. The black boundary indicates the neighbourhood for the Ams30k partition.}
\vspace{-5mm}
\label{fig:teaser}
\end{figure}

In Visual Place Recognition  the goal is to match a query image, for which the location is unknown, through Instance Search to a gallery of images with known geolocations. Application areas for this task include autonomous driving \cite{autonomous}, SLAM \cite{slam}, and matching historical architectural images to modern images \cite{gemert}. Research to develop VPR models has focused on several directions ranging from recognizing popular landmarks \cite{paris, delf, oxford} to arbitrary street scene localization with street-view images \cite{netvlad, 247, pittsburgh}. Until now place recognition methods have been exclusively developed with outdoor images as queries, yet often times indoor images provide information about the outside world through a window. Current methods fail to localize these images. We introduce the new task \textit{Inside Out Visual Place Recognition} (IOVPR). The goal for this task is to match the outdoor content that is visible in an indoor query image to a gallery of outdoor images, making it possible to localize images taken in an indoor setting. Figure \ref{fig:teaser}(a) shows such indoor queries with their ground truth location.

IOVPR can be employed for a wide range of applications, such as automated tagging of user-images on e.g. travel platforms or social media. More specifically it is useful for forensics, where localizing crime scenes is of great importance. Photographic evidence of indoor environments is common, since these criminal activities usually take place behind closed doors. In some cases, part of the outdoor environment is visible through a window. Organizations such as Bellingcat and Homeland Security Investigations have been involved in cases where images taken in an indoor environment as a house or car that contain a window have contributed to solving the case \cite{bellingcat, sunflower}. The amount of available relevant indoor images in a case varies, but as shown in the Sunflower Case \cite{sunflower}, one suitable image can already make a difference. For human trafficking or child abuse human lives can be saved when images depicting victims are localized \cite{hotels50k}. Another forensic application for IOVPR is the detection of housing fraud, where house owners exceed legal rental restrictions. Since address information is only released to tenants, IOVPR can be used to localize publicly advertised images. Localizing these images takes a lot of manual effort and time that we aim to reduce by introducing IOVPR. To support the investigative efforts in this direction, which are not supported by regular Visual Place Recognition, we explicitly focus on indoor environments by releasing a new dataset \textit{Amsterdam-XXXL}, as well as a novel training protocol \textit{Inside Out Data Augmentation}, that makes it possible to train Instance Search models for IOVPR. We empirically show the benefits of this training protocol on a smaller scale and show the difficulty of this large-scale dataset for existing methods.

The dataset introduced in this paper, Amsterdam-XXXL, consists of a spatially dense collection of 6.4 million panoramic street-view images and user photos from indoor locations, as visualized in Figure \ref{fig:teaser}(b). This dataset is 80 times larger on city-level than widely used VPR datasets, to provide a full coverage of the central part of Amsterdam. A second important aspect is the domain shift between the queries and the gallery. We are the first to present a large-scale VPR dataset with such a domain shift.

Additionally, we propose a novel data augmentation technique to adapt VPR models such that they are suitable for this new task. Specifically, we propose Inside Out Data Augmentation which we use to overlay indoor scenes onto street-view images with known geo-locations, synthetically creating `indoor' query images with recognizable  outdoor content. In this work we explore whether this data augmentation scheme is suitable for synthesizing training data for focusing the attention of existing VPR models on the window content. By using synthesized content it is possible to avoid the costly data collection and annotation process.

Our contributions are the following. (\textit{i}) We introduce \textit{Inside Out Visual Place Recognition} as a new task to localize indoor images through Instance Search. (\textit{ii}) We release a new dataset, \textit{Amsterdam-XXXL}, to encourage the development of new and improved techniques for IOVPR. (\textit{iii}) We propose \textit{Inside Out Data Augmentation} to train VPR models on synthetically generated indoor images and present baseline results to show the potential of our method and challenges on the large-scale aspects of this task.

\vspace{-1mm}
\section{Related Work}

\paragraph{Datasets.} Popular datasets that accelerated the development of Visual Place Recognition models for landmarks are Oxford Buildings \cite{oxford}, Paris \cite{paris-zisserman} and Holidays \cite{holidays}. Each of these datasets consists of a few thousand images and were suitable for early retrieval techniques because of their size and large number of positives for each query \cite{revox}. Deep Learning techniques required larger datasets which resulted in an extension of the Oxford and Paris datasets by the use of 1M distractor images \cite{revox}. Other landmark datasets that have been created for retrieval purposes are the Google Landmarks Dataset v1 \cite{delf} with 1M images and 15k distinct landmarks and its second version with over 5M images and over 200k distinct landmarks \cite{GLDv2}. Another fruitful direction in VPR research has been related to the use of (panoramic) street-view images. Starting from NetVLAD\cite{netvlad}, street-view datasets such as Pittsburgh 250k and Tokyo 24/7 use large amounts of images depicting the same places over time with the help of the Google Street View Time Machine. The Mapillary Street-level Sequences dataset is a more diverse dataset for VPR, with 1.6 million images from different cameras, captured over different times \cite{mapillary}. With data from different cities and continents, it aims to create models that generalize better to new environments. A recent variation of VPR is ground-to-aerial image geo-localization, where ground-view query images are compared to aerial image in a reference set \cite{ground, cvm, ground2}. Popular datasets for this task, e.g. CVUSA \cite{wide} and Vo and Hays' dataset \cite{vo}, use street-view images as query images.

\vspace{-3mm}
\paragraph{Methods.}
\label{sec:rw-method}
For VPR, we distinguish supervised and weakly-supervised methods. Supervised methods rely on a strong ground truth label, such as the class label for a landmark, to perform the optimization. These methods typically use local-feature based descriptors \cite{vlad, hqe}, global descriptors \cite{resnet, vgg}, or both \cite{delg, delf, revox}. 
Differently, when dealing with street-view images, the geolocation is used to select nearby images, but within this subset there is no ground-truth for visual matches. Therefore it is necessary to employ weakly-supervised methods. A common aspect of weakly-supervised models for VPR is the use of a triplet-based ranking loss. 

As one of the first methods that used triplet-based ranking for VPR, NetVLAD \cite{netvlad} uses a learnable layer based on VLAD \cite{vlad} to aggregate local descriptors and assign them to cluster centers. Extensions of NetVLAD are SARE \cite{sare}, CRN \cite{crn}, SFRS \cite{sfrs}, and Patch-NetVLAD \cite{patchnetvlad}. SARE \cite{sare} adds additional constrains on intra-place and inter-place feature embeddings, and CRN \cite{crn} uses context-aware feature reweighting to select the most relevant contexts for localization through efficient hard negative mining. SFRS \cite{sfrs} uses the NetVLAD backbone with self-supervised image-to-region similarities. These similarities return soft labels that further help the system find difficult positive images. This can be seen as additional, weak, training supervision which could help in refining fine-grained similarities. Patch-NetVLAD \cite{patchnetvlad} is a patch-based local descriptor method that reranks the top-100 results retrieved by NetVLAD. Our work adapts existing VPR models, making them suitable for the localization of indoor images, demonstrating the potential of IOVPR. 

\vspace{-1mm}
\section{Inside Out Visual Place Recognition}

While Visual Place Recognition aims to recognize the location of a given outdoor query image by finding the same location in a large gallery of outdoor images, we propose to modify this task to enable the localization of indoor images. Specifically, for Inside Out Visual Place Recognition the query set consists of indoor images and the gallery of outdoor images. The challenge thus becomes to identify the portions of the indoor image that contain information that can be used to match it to an outdoor image, and to perform matching based on partial and occluded visual information.

\subsection{Problem Formulation}
To enable image matching, Visual Place Recognition models are trained by feeding triplets consisting of a query image $q$, the easiest positive gallery image $p^{*}$ and a set of the most difficult negative gallery images $\{n_{j}|j\in\{1,...,N\}\}$ \cite{netvlad, sfrs, sare}. The easiest positive image $p^{*}$ is obtained by ranking all gallery images within a distance of 10 meters from query $q$ and taking the one with the highest visual similarity. The set of negatives is constructed by selecting the $N$ hardest negatives from a pool of 1000 randomly sampled negatives, where $N$ is usually set to 10. While all negative images are at least 25 meters away from the query, the hardest negatives are selected based on their visual similarity to the query.

For each training triplet $\{q,p^{*},\{n_{j}\}\}$, the goal is to learn a mapping function $d_{\theta}$, such that the distance $d_{\theta}(q,p^{*})$ between the training query $q$ and the easiest positive $p^{*}$ will be smaller than the distance $d_{\theta}(q,n_{j})$ between the query $q$ and all negative images $n_{j}$:
\begin{equation}
\label{eqn:dist}
d_{\theta}(q,p^{*}) < d_{\theta}(q,n_{j}), \forall j.
\end{equation}
Subsequently, the triplets of queries, positives, and negatives are used in a weakly-supervised ranking loss $L_{\theta}$:

\begin{equation}
\label{eqn:loss}
L_{\theta} = \sum_{j} l \Big( d_{\theta}^{2}(q,p^{*}) +m - d_{\theta}^{2}(q,n_{j})\Big) 
\end{equation}
where $l$ is the hinge loss $l(x) = \max(x, 0)$ and $m$ is the margin.

Considering the Inside Out Visual Place Recognition task, most query images are indoor images where the window only covers a small part of the image. This means that the majority of the visual content of an indoor image cannot be used to identify its location, as it does not match any of the outdoor gallery images. Because of this, the largest part of the image can be considered noise, especially for a model that has never learned to distinguish information about the indoor scene from the outdoor scene this can lead to a deterioration in performance. To measure the potential of existing models for Inside Out Visual Place Recognition, we introduce Inside Out Data Augmentation $M$ as a training protocol to enforce that models learn to differentiate between indoor and outdoor environments. $M$ will be used to augment the query only, with the help of a layout of an indoor space $c$ and a binary mask $b$. A detailed explanation of $M$, $b$, and $c$ will be given in Section \ref{sec:training_protocol}.

The goal for Inside Out Visual Place Recognition is to learn a representation in which the augmented query is closer to the easiest positive than it is to all negative images:

\begin{equation}
\label{eqn:dist_new}
d_{\theta}(M(q,b,c),p^{*}) < d_{\theta}(M(q,b,c),n_{j}), \forall j.
\end{equation}
To optimize for this new goal we modify the loss function in equation \ref{eqn:loss} and we formulate our new loss function as follows:

\begin{equation}
\label{eqn:loss_new}
L_{\theta} = \sum_{j} l \Big( d_{\theta}^{2}(M(q,b,c),p^{*}) +m - d_{\theta}^{2}(M(q,b,c),n_{j})\Big).
\end{equation}
Ideally we would have a large set of annotated indoor images to measure the potential of existing models for Inside Out Visual Place Recognition. However, such a dataset is currently not available and collecting it would be a costly process, since it should have a large variety of outdoor scenes that are visible through the window and this can only be manually determined. Our training protocol is introduced to overcome this barrier.

\begin{figure*}[t!]
\begin{center}
   \includegraphics[width=0.99\linewidth]{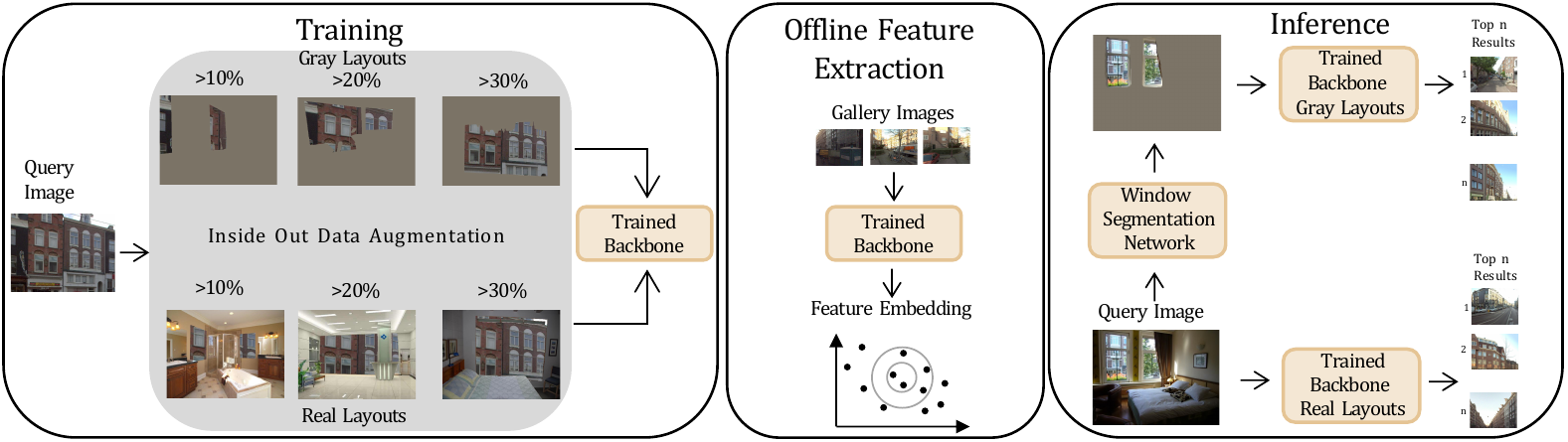}
\end{center}
\vspace{-3mm}
   \caption{Our approach has three phases: training, offline feature extraction, and inference. For training, we use Inside Out Data Augmentation to create synthetic training data by overlaying queries with gray and real layouts with window proportions of $>$10\%, $>$20\%, and $>$30\%. For each combination of layout type and proportion, we train three backbone models: NetVLAD, SARE, and SFRS. The training phase is followed by an offline feature extraction phase where the embeddings for the gallery images are calculated. At inference time, we distinguish between two scenarios: 1) the query is sent directly to the trained backbone for real layouts or 2) the query is first processed by a window segmentation network to predict its gray layout after which it is sent to the trained backbone for gray layout. In both scenarios, we obtain the top $n$ results of most similar gallery images to the queries.}
\vspace{-3mm}
\label{fig:method}
\end{figure*}
\vspace{-3mm}
\subsection{Inside Out Data Augmentation}
\label{sec:training_protocol}
The training protocol we propose, Inside Out Data Augmentation, consists of applying an augmentation function $M$ to each query image $q$ to create a synthetic indoor environment with an outdoor view. As input the augmentation function takes a query street-view image $q$, a layout $c$ of an indoor space with a window, and a binary mask $b$. The layout and binary mask are obtained from a reference dataset for semantic segmentation with pixel level annotations for windows. The layout represents the indoor scene itself and the binary mask is retrieved from applying a binary encoder on the layout that segments the layout into \textit{window} and \textit{non-window} categories. To perform the augmentation, we do a pixelwise multiplication between the street-view query image and the binary mask and a pixelwise multiplication between the layout and the inverse mapping of the binary mask. By adding these two components, we obtain a new query image with an indoor layout on top of it. The augmentation function is formalized as:
\begin{equation}
\label{eqn:mask}
M(q,b,c) = q \odot b + c \odot b^{-1}.
\end{equation}
For the layout, we consider two options: a real layout and a gray layout. The real layout refers to the original indoor scene of the reference image. The gray layout uses the same reference image, but every non-window pixel is matched to the mean color of the ImageNet training set to fully discard any irrelevant visual content. Examples of both layout types can be found in the Supplementary Material A2. During inference the masks to perform gray layout data augmentation are not available, so instead we use a segmentation network to detect the window portions of the query image. An overview of our proposed method is given in Figure~\ref{fig:method}, highlighting three stages: training, offline feature extraction and inference.

\vspace{-1mm}
\section{The Amsterdam-XXXL Dataset}
To evaluate the performance of models on IOVPR we present the Amsterdam-XXXL dataset. This dataset consists of three partitions: 1) Outdoor-Ams, 2) Indoor-Ams, and 3) Ams30k. The datapoints are presented in Figure \ref{fig:teaser}(b).\footnote{Map tiles by Stamen Design, under CC BY 3.0. Data by OpenStreetMap.} Outdoor-Ams and Ams30k consist of panoramic street-view images that are selected from a set that has been made publicly available by the Municipality of Amsterdam.\footnote{https://api.data.amsterdam.nl/} Indoor-Ams consists of two sets of 500 user-generated photos, one consisting of images with creative commons licenses and the other provided by TripAdvisor.\footnote{A dataset provided by © 2020 TripAdvisor LLC was used for analysis and visualisation.} Examples of these images for each data source are presented in the Supplementary Material A1. All partitions will be published and made available for research purposes.

\vspace{-3mm}
\paragraph{Outdoor-Ams.} This partition consists of 6,467,112 GPS-annotated street-view images taken at 269,463 locations in Amsterdam, where each panoramic image is cut into 24 partially overlapping images of $480 \times 640$. With 6.4M images, this is much larger on city-level than widely used VPR datasets and results in a more challenging dataset. Details regarding the coverage of the area and the cutting process are described in the Supplementary Material A3. 

\vspace{-3mm}
\paragraph{Indoor-Ams.} This partition consists of two query sets for evaluation. To test the performance of our approach properly, at least a few hundred images which show an outdoor scene through a window, but which are taken indoors, are needed. We construct the Indoor-Ams set for this purpose, which consists of a validation and test set of 500 indoor images each with a view of an outdoor scene. These splits are created based on the image license, ensuring consistent licensing within each split. The validation set consists of 500 images in total from platforms with images with Creative Commons licenses (335), namely Flickr, Unsplash, and Traffickcam and self-made pictures taken from public and university buildings(165). The test set consists of 500 images that are selected from an image collection of user images provided by TripAdvisor, consisting of hotels, restaurants, and other visiting areas in Amsterdam. 

Indoor-Ams consists of images taken in the time period from 2004 until 2021. We manually verified that all query images (even ones from before the panorama images were taken) can be visually matched to the appearance of the location in the panoramic images. About 5\% of these images are night images, where we made sure that the outdoor area is still visible. The window proportions vary from 10-80\%. Due to most neighborhoods being mainly residential, we decided not to take pictures there to respect the resident’s privacy. Therefore our dataset only contains public images from residential places with a Creative Commons license. As a consequence, the majority of the images are collected from and taken at public, commercial, and university spaces and released with a license that can be used for research purposes. Therefore Indoor-Ams has a spatial bias to the center of Amsterdam. However, this spatial bias should not affect the methods that are trained on Amsterdam-XXXL, since the query sets only serve as a validation and test set. 

The annotation process has been done by three annotators who live in Amsterdam. They manually inspected each image and its tag. Details about the selection and annotation process are described in the Supplementary Material A4.

\vspace{-3mm}
\paragraph{Ams30k.} This is created for training purposes and is for this reason modelled after Pitts30k \cite{netvlad}, where 30k represents the number of gallery images. The total number of panoramas in Ams30k is 2443 which results in almost 59k street-view images. Details about its splits can be found in the Table 1 in the Supplementary Material A5. To guarantee variation in viewpoint, light, and occlusion, the gallery images are from 2018 and the query images from 2019. We select the geographical region 'the Pijp' in Amsterdam as a region for this dataset with good coverage, since it has a large variety in architectural styles, a good balance of houses, restaurants and stores, and green areas (i.e.\ parks). We define the train, validation, and test set through geographical boundaries, enforcing that the three sets have a similar surface area. 
\vspace{-6mm}
\section{Experimental Setup}

\paragraph{Inside Out Data Augmentation.}
To perform Inside Out Data Augmentation as explained in Section \ref{sec:training_protocol}, we take a subset of ADE20K. This dataset has images that cover a range of scenes and objects and are pixel annotated with 150 different classes. Our subset consists of indoor images with a window, more specifically 4687 images in the train set and 461 images in its validation set. After reshaping these images to $480 \times 640$, we apply our layout function from Equation \ref{eqn:mask} and create two sets by using two layouts, the original layout and a gray layout. Examples of the output are shown in Supplementary Material A2. To measure the effect of the percentage of available window pixels for training and inference, we create three different sets of images per data augmentation type depending on the window proportion of the layouts, namely $>$10\%, $>$20\%, and $>$30\%. In this setup, $>k$\% stands for $k$\%-100\%.
\vspace{-3mm}

\paragraph{Backbones.}
We evaluate four model architectures as possible backbones: NetVLAD\cite{netvlad}, SARE \cite{sare}, SFRS \cite{sfrs}, and Patch-NetVLAD which were briefly discussed in Section \ref{sec:rw-method}. NetVLAD is a trainable VLAD layer, inspired by the Vector of Locally Aggregated Descriptors \cite{vlad}, a popular descriptor pooling method for instance retrieval before the Deep Learning era. It generates global features of images by aggregating feature maps by summing residuals between features and cluster centers. SARE \cite{sare} which stands for Stochastic Attraction-Repulsion Embedding aims to capture the intra-place attraction and inter-place repulsion in the embedding space. This is achieved by minimizing the KL divergence between the learned and the actual probability distributions. SFRS \cite{sfrs} refers to Self-supervising Fine-grained Region Similarities. It splits image feature maps in 4 half and 4 quarter regions and computes similarity scores between these regions and the query during training. These similarities are used to supervise the network through a soft cross-entropy loss on top of the triplet loss, to enhance the learning of local features. Patch-NetVLAD \cite{patchnetvlad} combines global with local descriptor methods. It uses NetVLAD descriptors to retrieve the top-$k$, with default $k=100$, most likely matches and performs local matching of patch-level descriptors at multiple scales to reorder this top-$k$. Then a RANSAC-based scoring method is used for higher retrieval performance. We test a new model architecture by combining the patch-based local feature method of Patch-NetVLAD with the SFRS backbone and show initial results on Outdoor-Ams. We name this method \textit{Patch-SFRS}. 

\vspace{-3mm}
\paragraph{Window Segmentation Network.}
\label{sec:wsn}

To obtain a gray layout at inference time for the gray layout model, we finetune the network UPerNet50 \cite{upernet} for window segmentation.  For finetuning we use the same subset of the ADE20K dataset as for Inside Out Data Augmentation \cite{ade20k}. After finetuning UPerNet50 on two classes (i.e., window and non-window) for three epochs we obtain an accuracy score of 95\% on the validation set of ADE20K. 

\vspace{-3mm}
\paragraph{Evaluation Criteria.}
As is typical for VPR we report our results using the Recall@K metric \cite{recall}. More specifically, the query image is seen as correctly localized if at least one of the top K retrieved gallery images is within a radius of $n$ meters from the ground truth position of the query. The radius for VPR is usually set to 25 meters. Since the street-view images can be taken further away from the building, we increase the distance to 50 meters for our experiments on Indoor-Ams. Regarding the recall, we set our K values for our small sets to 1, 5, 10, 15, 20, and 25. For our large set of 6.4M images, we add 50, 75, and 100.

\begin{figure*}[t!]
\begin{center}
   \includegraphics[width=0.9\linewidth]{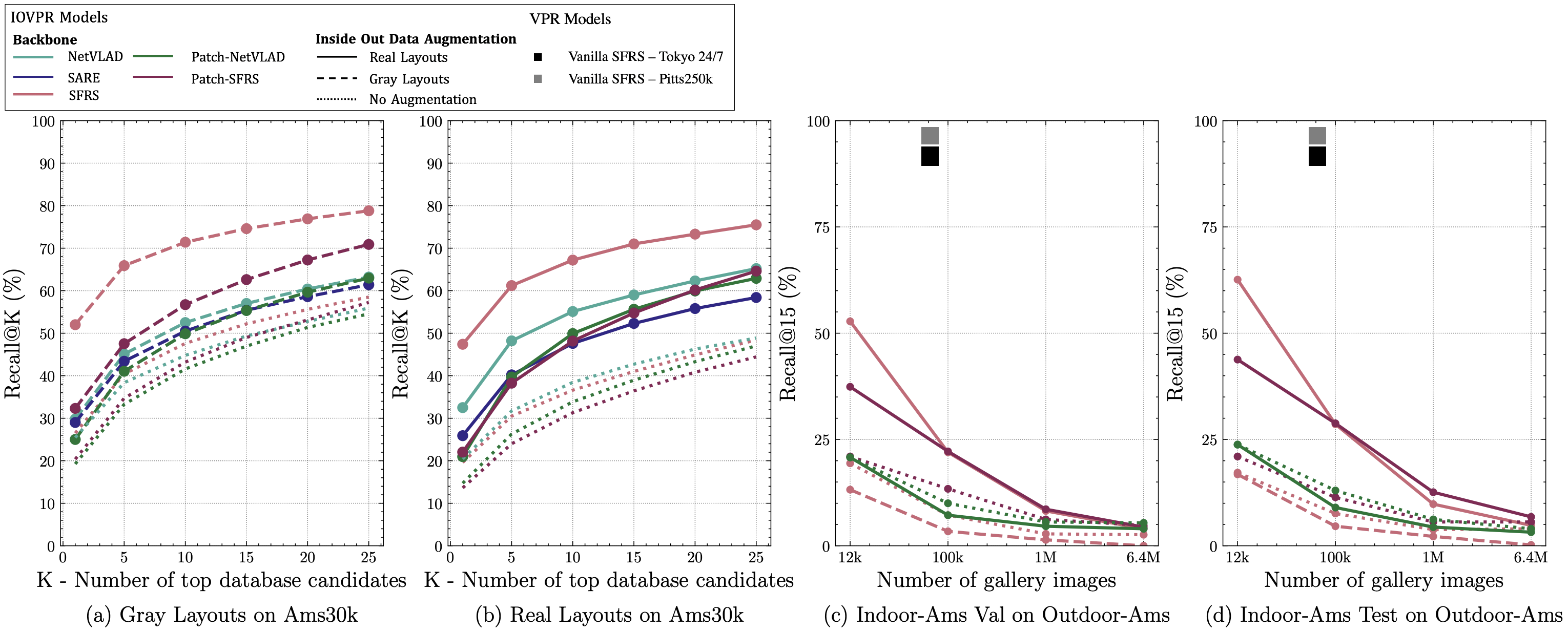}
\end{center}
\vspace{-3mm}
   \caption{Results of Inside Out Visual Place Recognition on the Amsterdam-XXXL dataset. Figure (a) and (b) present the results for gray and real layouts on Ams30k, where the dotted pink line indicates the results of not using Inside Out Data Augmentation (i.e., no layouts). (c) and (d) show the results on Indoor-Ams for the val and test set respectively, evaluated on subsets of Outdoor-Ams. This is best viewed in color.}
\vspace{-3mm}
\label{fig:results}
\end{figure*}
\vspace{-3mm}
\section{Results}

\paragraph{Inside Out Data Augmentation.} In our experiments, we analyze the effect of Inside Out Data Augmentation on our queries during training. In particular we study the difference between using gray and real layouts. In addition, we take models trained on Ams30k without layouts, to measure the contribution of Inside Out Data Augmentation. In Figure \ref{fig:results}, the results are presented on Ams30k for gray (a) and real layouts (b) with a window proportion of $>$20\%. The results for $>$10\% and $>$30\% and the corresponding tables can be found in Supplementary Material B Tables 2-3. The best performing model in both setups is SFRS with a window proportion of $>$20\%, where the model for gray layouts (66\% Recall@5) performs slightly better than for real layouts (61\% Recall@5). Comparing this to SFRS that has not been trained with Inside Out Data Augmentation, we notice a difference of 25\% with the best model for gray layouts and 31\% with real layouts. We  conclude that the use of Inside Out Data Augmentation significantly helps in localizing indoor images from Ams30k and that the benefits of the gray and real layouts are similar.
\vspace{-3mm}
\paragraph{Window Proportion.} We study the effect of the proportion of available window pixels in images by setting a lower bound on the amount of visible window pixels. For both the gray and the real layouts of Ams30k we compare three lower bounds, namely $>$10\%, $>$20\%, and $>$30\% of the full image. We study how well these different training setups generalize by training on these three different sets and testing on a test set with a window proportion of at least $>10\%$ of the full image. Considering the different window proportions, Figure B.1 in the Supplementary Material indicates that the $>$10\% proportion performs worst out of the three proportions. We expect that for this proportion there is insufficient information for models to learn about outdoor environments. We also see that for gray and real layouts, the models trained on a window proportion of $>$20\% and $>$30\% perform well on the test set, despite the test set containing images with a window proportion of $>$10\%. Overall the highest performance is obtained by models trained on images with a $>$20\% proportion. For this reason we recommend training on images with at least a decent window proportion of 20\% to make sure that the network has enough opportunities to optimize for this task.  
\vspace{-3mm}
\paragraph{Backbone Comparison.} For each of the chosen backbones NetVLAD, SARE, and SFRS we measure the performance on different layouts and window proportions by employing three different backbones, two types of layouts, and three window proportions. Additionally, we evaluate Patch-NetVLAD and Patch-SFRS on Ams30k queries with gray layouts and real layouts. With respect to the choice of backbone, we see that a strong model backbone contributes more to the performance than the window proportion that is used during training, except for when the model has not seen any layouts during training. In general we see that the SFRS model outperforms the other backbones in every setup. An unexpected result is that Patch-SFRS performs worse than SFRS on Ams30k for both real and gray layouts. When not using any layouts during training, Patch-SFRS does also not improve the performance. From Figure \ref{fig:results}(d) we see that Patch-SFRS performs worse than SFRS on a small scale subset. We expect that the higher overall performance of SFRS might be due to its fine-grained approach, which utilizes region-similarities, allowing SFRS to focus on the information rich window parts of the images. Hence we recommend future approaches to focus on fine-grained techniques. 

\begin{figure*}[t!]
\begin{center}
   \includegraphics[width=0.90\linewidth]{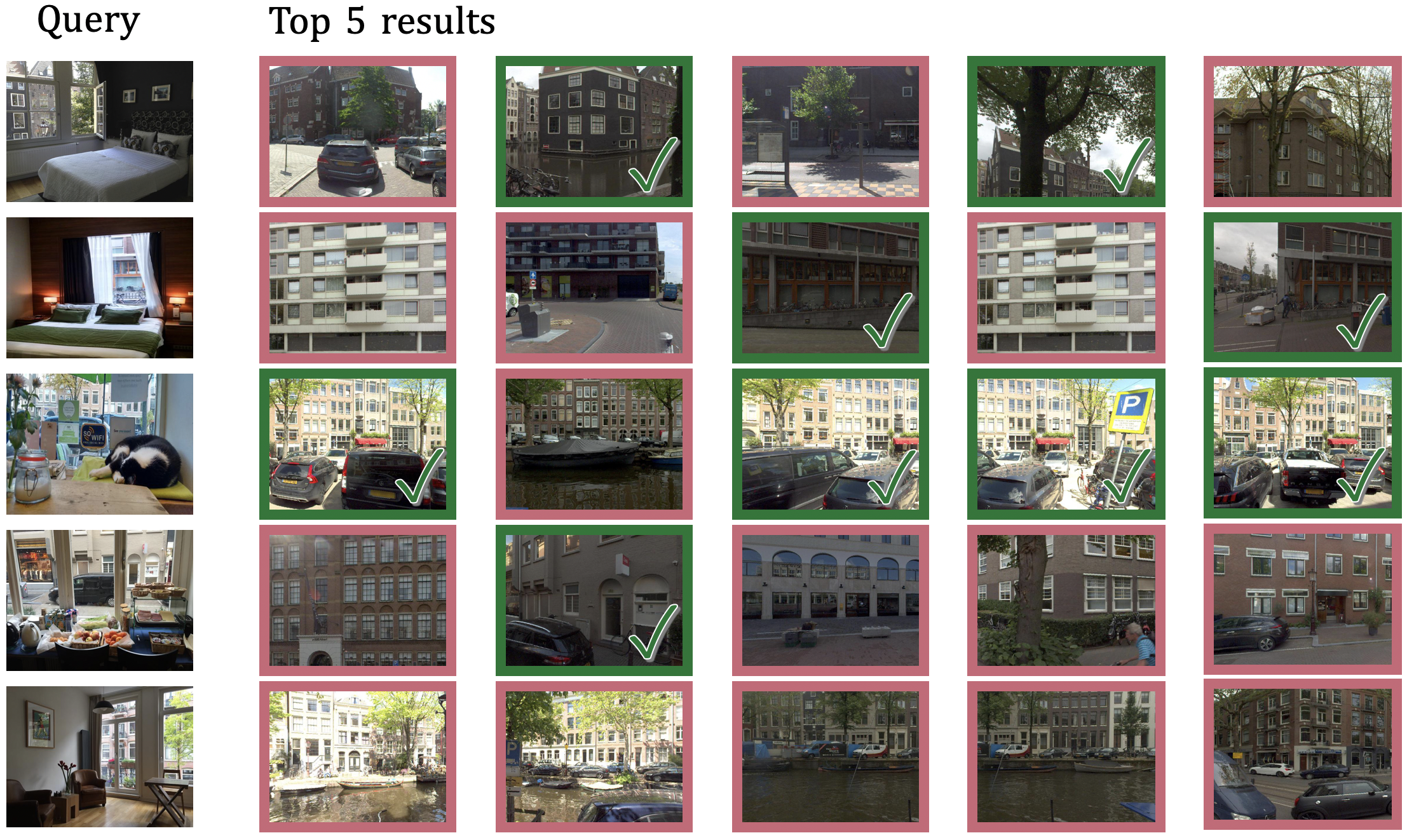}
\end{center}
\vspace{-3mm}
   \caption{Qualitative results of Inside Out Visual Place Recognition success and failure cases. These results demonstrate the potential of our approach in returning the correct location (sometimes even multiple times), but also how challenging it is to find the correct match when very little is visible of the outdoor scene.}
\vspace{-3mm}
\label{fig:q}
\end{figure*}
\vspace{-3mm}
\paragraph{Large Scale Visual Place Recognition.}
To get an understanding of how well our proposed approach scales to a large gallery, we evaluate the performance of using indoor images captured by users. Some success and failure cases from our best performing model on the test set of Indoor-Ams are presented in Figure \ref{fig:q}. The difficulty of VPR is directly related to the number of gallery images, due to the potential for false positives (i.e. distractor images being ranked high) increasing
with gallery size. With 6.4M images Amsterdam-XXXL is 80 times larger than widely used VPR datasets on city-level (e.g., Tokyo 24/7, Pitts250k, and city-splits of Mapillary Street-level Sequences Dataset) and thus includes many more distractors. To show the effect of distractors we zoom in on the reported results by creating three subsets of Outdoor-Ams of 12k, 100k, and 1M images. In Figure \ref{fig:results}(c) we present the Recall@15 results of SFRS, Patch-NetVLAD, and Patch-SFRS trained with and without our data augmentation methods of 20\% for the validation set of Indoor-Ams, Figure \ref{fig:results}(d) shows the same results for the test-set. As demonstrated in this figure, when evaluated with fewer distractors (as is typical for VPR) SFRS with real layouts shows a large improvement over the SOTA of up to 50\% and 30\% for the 12k and 100k subsets. However, if the number of distractors is increased to well beyond what is common for standard VPR, the performance for both methods collapses. For a high number of distractors we see that reranking the top-100 results improves the performance and therefore Patch-SFRS performs best.

To highlight the differences between IOVPR and VPR, we add a black and gray box, for vanilla SFRS models trained on Tokyo 24/7 and Pitts250k respectively as reported in \cite{sfrs}, to Figure \ref{fig:results}. On VPR this model almost achieves 100\% correctness. The same model has a score of nearly 8\% on a similarly sized subset of Outdoor-Ams. Although our proposed modifications boost this to almost 30\%, existing VPR methods are clearly not suited for IOVPR. Given these results we conclude that the scale of Outdoor-Ams remains an interesting challenge for Visual Place Recognition.

\vspace{-3mm}
\section{Conclusion}
In this work, we have introduced the novel task Inside Out Visual Place Recognition and a large-scale dataset Amsterdam-XXXL to train and evaluate this task on. Additionally, we proposed Inside Out Data Augmentation as a new training protocol to extend Visual Place Recognition for localizing indoor images. The results on Ams30k and subsets of Outdoor-Ams show the potential of Inside Out Data Augmentation, on both synthetic and real indoor images. However, this task remains a challenge for existing Visual-only Place Recognition methods on a large-scale dataset with millions of images. This opens up a range of possible future directions, including multimodal approaches.

We are aware of the fact that techniques for Inside Out Visual Place Recognition have the potential to cause harm related to the privacy of individuals when misused. While we condemn any non-ethical applications, we hope to inspire the research community to work on this task for good purposes and to help advance the state-of-the-art for Visual Place Recognition on indoor environments.

\section*{Acknowledgements}
We thank Jos de Winde and TripAdvisor for their help in creating the dataset. This research was supported by the Nationale Politie. All content represents the opinion of the authors, which is not necessarily shared or endorsed by their respective employers and/or sponsors.

\let\url\nolinkurl
\bibliography{main}
\clearpage
\appendix
\counterwithin{figure}{section}
\section{Amsterdam-XXXL}

\subsection{Images Indoor-Ams}

\begin{figure}[h]
\begin{center}
   \includegraphics[width=0.99\linewidth]{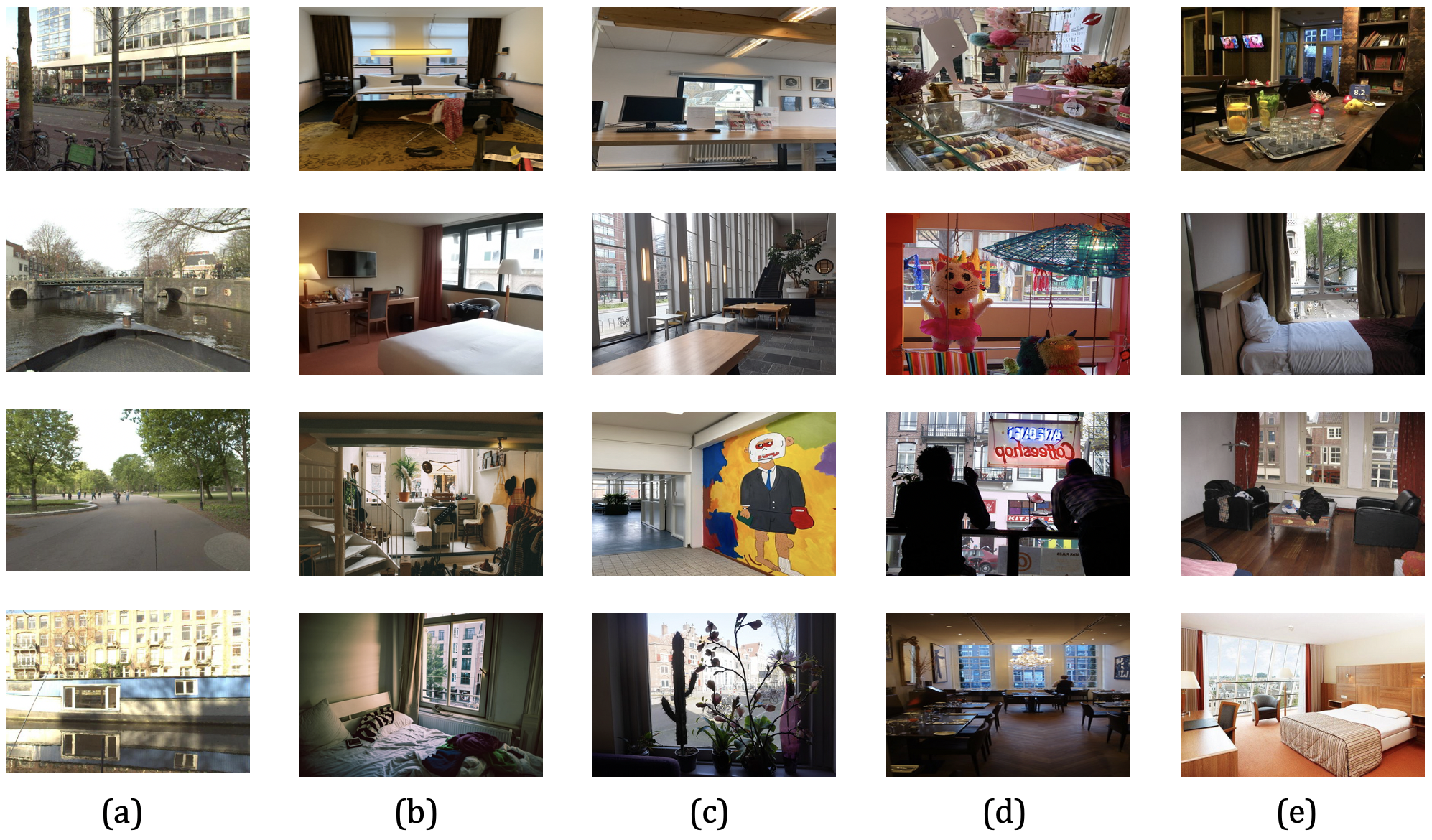}
\end{center}
   \caption{Examples of images from Amsterdam-XXXL. (a) processed street-view images, used for Outdoor-Ams and Ams30k. (b), (c) and (d) are images from the validation set of Indoor-Ams, where (b) has images of Traffickcam (upper two) and Unsplash (bottom two), (c) our selfmade pictures taken from public buildings, and (d) images from Flickr. (e) consists of images from the test set of Indoor-Ams, these are user images from TripAdvisor.}
\label{fig:1}
\end{figure}

Figure \ref{fig:1} presents images from the several data sources in Amsterdam-XXL.

\subsection{Data Augmentation on Ams30k}

\begin{figure}[h]
\begin{center}
   \includegraphics[width=0.99\linewidth]{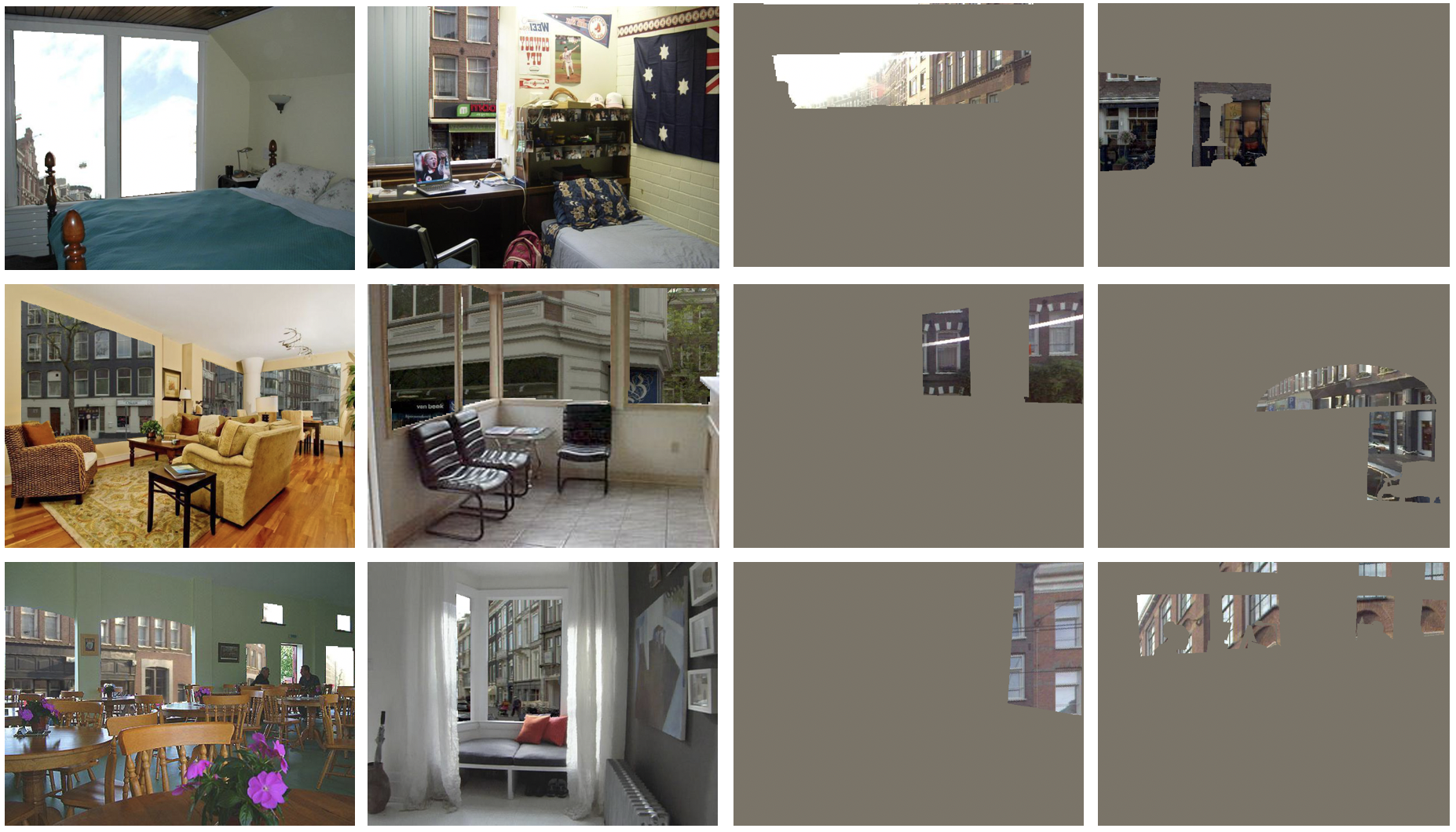}
\end{center}
   \caption{Examples of query images of Ams30k with real and gray layouts that are used during training}
\label{fig:aug}
\end{figure}

Figure \ref{fig:aug} presents the result of our data augmentation technique. These images are used as queries during training on Ams30k.

\subsection{Outdoor-Ams}

For this collection we use images created in 2018, 2019, and 2020, and create the set in the following manner. First, we retrieve the metadata of all images taken in these years from the API of the Municipality of Amsterdam. With the help of shapefiles that indicate the boundaries of regions in Amsterdam\footnote{https://data.lab.fiware.org/dataset/gebiedsindeling\_amsterdam}, we select only the images that are part of central Amsterdam, that is surrounded by a highway. Then we use the DBScan algorithm \cite{dbscan} to select a set of panoramas such that at least every 5 meters of the selected area is covered by a panorama from our set. For this, we use PostGIS, a spatial database extender for PostgreSQL, and more specifically ST\_ClusterDBSCAN. This option creates clusters of points with less than 5 meters distance. Note that the quality of the overall coverage depends on where the vehicle has passed for its recordings, but apart from some pedestrian streets in the central area of Amsterdam, this coverage is nearly complete. The panoramas we analyze for this set have resolution $2000 \times 4000$ and are processed similarly as was done in \cite{netvlad}. Firstly, the image is processed to exclude irrelevant content, such as the sky, the ground, and the vehicle that collected the images. This is done using an equilateral projection by which the panoramas are projected into six images of $960 \times 960$: front, left, right, back, top, and bottom. As the top and bottom images only capture sky and ground respectively they are excluded. Subsequently the front, back, left, and right images are stitched together to create one image of $960 \times 3840$ pixels. From this stitched image the bottom $240$ pixels are then removed to remove the area that captured the vehicle, resulting in a single panoramic image of $720 \times 3840$ that only captures the surroundings of the vehicle which might be used for Visual Place Recognition. To produce the perspective images that make up the Outdoor-Ams set, this resulting image is cut into 24 partially overlapping images of $480 \times 640$ for two pitch levels and twelve different yaws. Note that there is an overlap of $240$ vertical and $320$ horizontal pixels between the perspective images to increase the likelihood of objects being fully visible, and not split by the boundaries between images.

\subsection{Indoor-Ams}
The validation set of Indoor-Ams is created collecting images with a suitable license from websites such as Flickr. Since the majority of the images does not contain GPS coordinates in the metadata, we focused on searching by keywords. We searched for generic keywords such as 'restaurant', 'cafe', 'hotel', 'coffee', 'beer' combined with 'Amsterdam'. But also more specific keywords, such as names of streets, hotels, restaurants and shops. From the retrieved results, we manually selected the images with windows and added only directly the images to our dataset that had an exact address or a store name. For images with only a neighborhood indication, we manually tried to verify the location with the help of Google Maps and added the image to our dataset whenever we had a match with a nearby panorama image in Outdoor-Ams.

\begin{figure}[h!]
\begin{center}
   \includegraphics[width=0.99\linewidth]{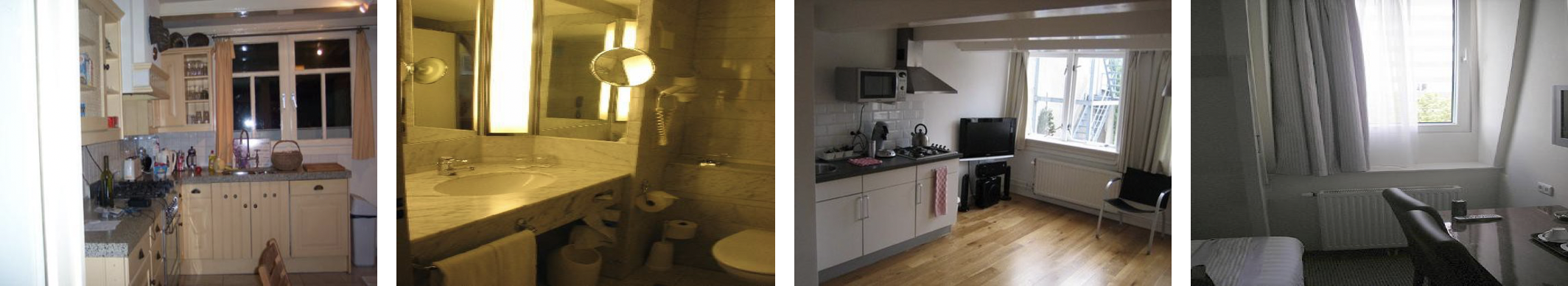}
\end{center}
   \caption{Examples of images not usable for for Inside Out Visual Place Recognition because they contain incorrectly detected windows, views which are "non-detectable", or outdoor scenes with extreme exposure (i.e., underexposure due to nighttime, or overexposure due to bright sunlight.}
\label{fig:not_suitable}
\end{figure}

For the test set of Indoor-Ams we used images of TripAdvisor, which were all tagged with a location name. With the help of the Google Maps API, we restricted this set to images in the area from Outdoor-Ams. Consecutively, a window segmentation network, which is explained in Section 5, is used to detect images with at least 5\% of their pixels labeled as window. From these remaining images we manually selected 500 images which were most suitable for the task and which had a nearby panorama image in the Outdoor-Ams set. During the creation process of the Indoor-Ams test set, multiple types of images were detected by the window segmentation network that are not suitable for Inside Out Visual Place Recognition. Examples of such images are shown in Figure \ref{fig:not_suitable}.

\subsection{Ams30k}

In Table \ref{table:splits-app}, the setup of the Ams30k partition is presented. For each split, the number of unique locations is equal to the number of panoramic street-view images. By processing these street-view images to 24 images of size $480 \times 640$ each, we obtain the total number of images.

\begin{table}[h!]
\centering
\begin{tabular}{ |c|c|c| } 
\hline
\textbf{Subset} & \textbf{\# Unique locations} & \textbf{\# Images} \\
\hline
Train query & 417 & 10008 \\
Train gallery & 466 & 11184 \\
Val query & 366 & 8784 \\
Val gallery & 385 & 9240 \\
Test query & 387 & 9288 \\
Test gallery & 422 & 10128 \\
\hline
\end{tabular}
\caption{Statistics of Ams30k}
\label{table:splits-app}
\end{table}

\section{Results}

\begin{figure*}[ht!]
\begin{center}
   \includegraphics[width=0.8\linewidth]{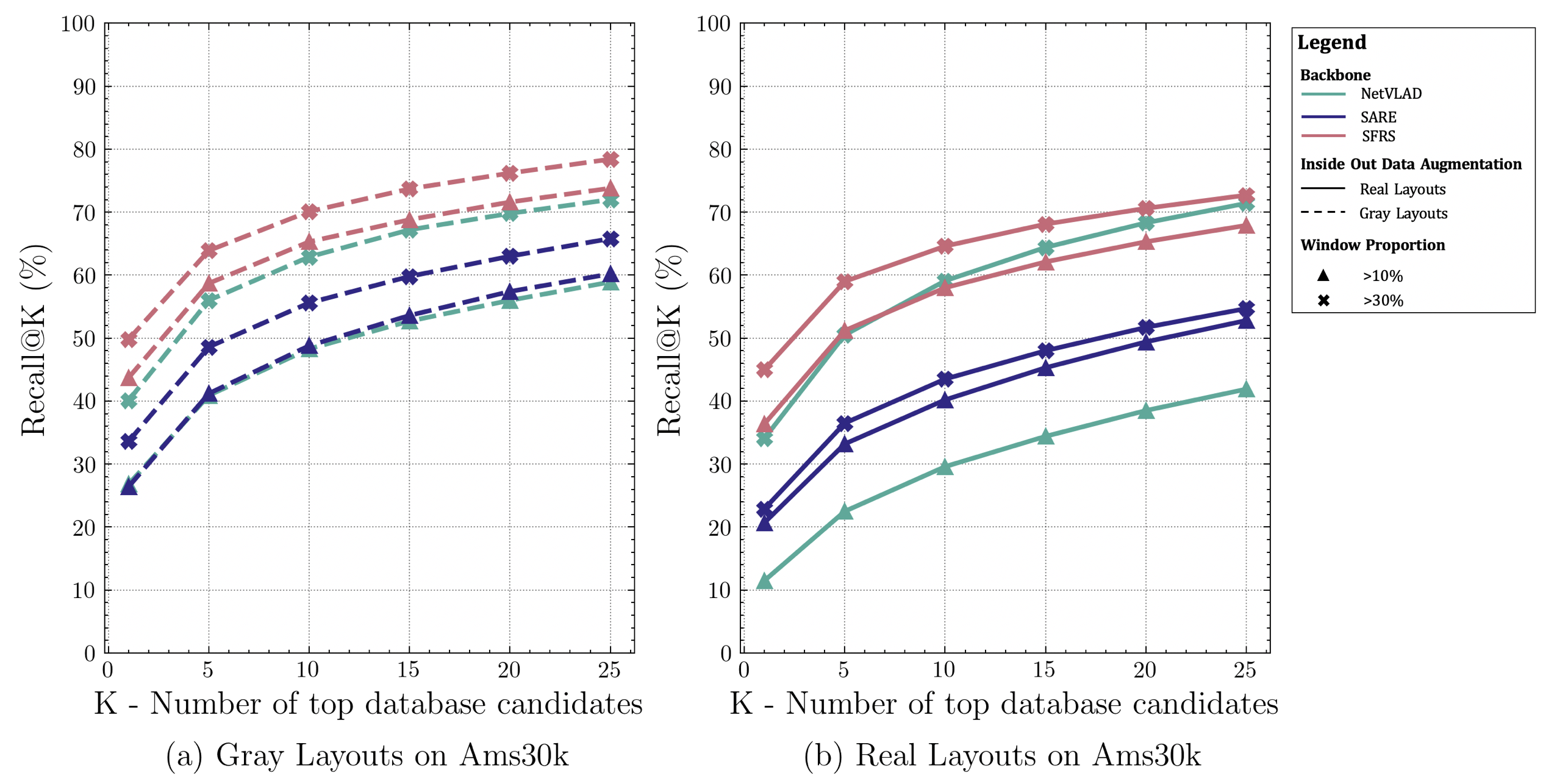}
\end{center}
\vspace{-3mm}
   \caption{Results of Inside Out Visual Place Recognition on the Ams30k. Figure (a) and (b) present the results for gray and real layouts on Ams30k for 10\% and 30\%. This is best viewed in color.}
\vspace{-3mm}
\label{fig:results_appendix}
\end{figure*}

Figure \ref{fig:results_appendix} presents the results corresponding to Figure 3(a) and 3(b) in the main paper, but for window ratios of $>$10\% and $>$30\%.

Table \ref{table:4a}-\ref{table:4d} correspond to Figure 3(a) - 3(d) from the main paper respectively.   

\begin{table*}[h!]
\centering
\begin{tabular}{ |c|c|c|c|c|c|c| } 
\hline
\textbf{Model} & \textbf{R@1} & \textbf{R@5} & \textbf{R@10} & \textbf{R@15} & \textbf{R@20} & \textbf{R@25} \\
\hline
NetVLAD $>$10\% & 26.8 & 40.9 & 48.3 & 52.7 & 56.0 & 58.9 \\ 
NetVLAD $>$20\% & 29.8 & 45.1 & 52.5 & 57.0 & 60.4 & 63.2 \\
NetVLAD $>$30\% & 40.1 & 56.0 & 62.9 & 67.2 & 69.8 & 72.0 \\ 
SARE $>$10\% & 26.4 & 41.2 & 48.8 & 53.6 & 57.4 & 60.2 \\
SARE $>$20\% & 29.0 & 43.4 & 50.5 & 55.4 & 58.6 & 61.4 \\ 
SARE $>$30\% & 33.6 & 48.6 & 55.6 & 59.8 & 63.0 & 65.8 \\ 
SFRS $>$10\% & 43.7 & 58.7 & 65.3 & 68.8 & 71.6 & 73.8 \\ 
SFRS $>$20\% & \textbf{52.0} & \textbf{65.9} & \textbf{71.4} & \textbf{74.6} & \textbf{76.9} & \textbf{78.8} \\ 
SFRS $>$30\% & 49.8 & 63.9 & 70.1 & 73.7 & 76.2 & 78.4 \\ 
NetVLAD - No Augmentation & 25.5 & 38.4 & 44.8 & 49.3 & 52.8 & 55.9 \\
Patch-NetVLAD - No Augmentation & 19.3 & 33.3 & 41.6 & 47.0 & 51.4 & 54.5 \\
Patch-NetVLAD $>$20\% & 25.0 & 41.1 & 49.9 & 55.3 & 59.7 & 63.0 \\
SFRS - No Augmentation & 26.5 & 40.4 & 47.6 & 52.2 & 55.6 & 58.5 \\
Patch-SFRS - No Augmentation & 20.4 & 34.7 & 43.2 & 49.0 & 53.1 & 57.2 \\
Patch-SFRS $>$20\% & 32.3 & 47.5 & 56.7 & 62.6 & 67.2 & 70.9 \\

\hline
\end{tabular}
\caption{Results for Inside Out Data Augmentation with gray layouts on Ams30k, corresponding to Fig 3(a) and B.1(a)}
\label{table:4a}
\end{table*}

\begin{table*}[h!]
\centering
\begin{tabular}{ |c|c|c|c|c|c|c| } 
\hline
\textbf{Model} & \textbf{R@1} & \textbf{R@5} & \textbf{R@10} & \textbf{R@15} & \textbf{R@20} & \textbf{R@25} \\
\hline
NetVLAD $>$10\% & 11.5 & 22.5 & 29.6 & 34.4 & 38.5 & 41.9 \\
NetVLAD $>$20\% & 32.5 & 48.2 & 55.1 & 59.0 & 62.3 & 65.2 \\ 
NetVLAD $>$30\% & 34.1 & 50.5 & 59.1 & 64.4 & 68.3 & 71.4 \\ 
SARE $>$10\% & 20.7 & 33.2 & 40.2 & 45.3 & 49.4 & 52.8 \\ 
SARE $>$20\% & 25.9 & 40.2 & 47.6 & 52.3 & 55.8 & 58.4 \\ 
SARE $>$30\% & 22.9 & 36.5 & 43.5 & 48.0 & 51.7 & 54.7 \\ 
SFRS $>$10\% & 36.4 & 51.2 & 58.0 & 62.1 & 65.3 & 67.9 \\ 
SFRS $>$20\% & \textbf{47.4} & \textbf{61.2} & \textbf{67.2} & \textbf{71.0} & \textbf{73.3} & \textbf{75.5} \\ 
SFRS $>$30\% & 45.0 & 59.0 & 64.6 & 68.1 & 70.6 & 72.7 \\ 
NetVLAD - No Augmentation & 20.4 & 31.7 & 38.4 & 42.7 & 46.3 & 48.9 \\
Patch-NetVLAD - No Augmentation & 14.7 & 26.3 & 33.9 & 39.0 & 43.3 & 47.0 \\
Patch-NetVLAD $>$20\% & 21.0 & 39.7 & 49.9 & 55.6 & 59.9 & 62.9 \\
SFRS - No Augmentation & 19.6 & 30.5 & 36.6 & 41.0 & 44.9 & 48.4 \\
Patch-SFRS - No Augmentation & 13.6 & 24.0 & 31.3 & 36.4 & 40.8 & 44.4 \\
Patch-SFRS $>$20\% & 22.0 & 38.3 & 48.2 & 54.8 & 60.2 & 64.6 \\

\hline
\end{tabular}
\caption{Results for Inside Out Data Augmentation with real layouts on Ams30k, corresponding to Fig 3(b) and B.1(b)}
\label{table:4b}
\end{table*}

\begin{table*}[h!]
\centering
\resizebox{\textwidth}{!}{\begin{tabular}{ |c|c|c|c|c|c|c|c|c|c|c| }
\hline
\textbf{Model} & \textbf{Size} & \textbf{R@1} & \textbf{R@5} & \textbf{R@10} & \textbf{R@15} & \textbf{R@20} & \textbf{R@25} & \textbf{R@50} & \textbf{R@75} & \textbf{R@100}\\
\hline
SFRS - No Augmentation & 12k &  4.4 & 11.0 & 16.8 & 19.4 & 22.2 & 25.8 & 36.2 & 41.6 & 49.2 \\ 
SFRS - Gray Layouts & 12k & 2.2 & 7.6 & 9.8 & 13.2 & 16.0 & 18.4 & 27.6 & 37.0 & 44.6 \\ 
SFRS - Real Layouts & 12k & \textbf{12.2} & \textbf{32.0} & \textbf{43.0} & \textbf{52.8} & \textbf{59.0} & \textbf{63.4} & \textbf{80.0} & \textbf{88.4} & \textbf{92.4}\\
Patch-SFRS - No Augmentation & 12k &  5.4 & 12.6 & 18.0 & 21.0 & 23.8 & 26.6 & 37.6 & 44.6 & 49.2\\
Patch-SFRS - Real Layouts & 12k & 7.6 & 19.8 & 30.0 & 37.4 & 44.2 & 50.8 & 71.6 & 84.6 & 92.4\\
Patch-NetVLAD - No Augmentation & 12k & 5.2 & 12.0 & 17.2 & 20.8 & 23.6 & 27.0 & 37.4 & 44.4 & 49.8\\
Patch-NetVLAD - Real Layouts & 12k & 5.4 & 10.8 & 16.2 & 20.8 & 24.0 & 26.4 & 40.0 & 46.8 & 49.6\\
\hline
SFRS - No Augmentation & 100k &  1.8 & 4.8 & 6.0 & 7.2 & 8.4 & 10.8 & 16.2 & 19.6 & 24.0 \\ 
SFRS - Gray Layouts & 100k & 0.8 & 1.4 & 2.4 & 3.4 & 3.8 & 4.8 & 7.0 & 9.4 & 10.8 \\ 
SFRS - Real Layouts & 100k & \textbf{6.0} & \textbf{11.2} & \textbf{17.4} & \textbf{22.0} & \textbf{24.0} & \textbf{27.6} & \textbf{37.6} & \textbf{43.2} & \textbf{49.2}\\
Patch-SFRS - No Augmentation & 100k & 4.8 & 7.6 & 11.4 & 13.4 & 14.8 & 15.8 & 20.6 & 22.8 & 24.0\\
Patch-SFRS - Real Layouts & 100k & 4.2 & 12.2 & 17.0 & 22.2 & 25.0 & 28.4 & 38.8 & 44.8 & 49.2\\
Patch-NetVLAD - No Augmentation & 100k & 2.8 & 6.0 & 8.2 & 10.0 & 12.4 & 14.8 & 20.8 & 23.0 & 25.0\\
Patch-NetVLAD - Real Layouts & 100k & 3.4 & 5.0 & 5.8 & 7.2 & 9.0 & 10.2 & 14.6 & 16.6 & 17.2\\
\hline
SFRS - No Augmentation & 1M & 1.2 & 2.0 & 2.4 & 2.8 & 3.6 & 4.4 & 7.2 & 8.2 & 9.8 \\ 
SFRS - Gray Layouts & 1M & 0.2 & 0.4 & 0.6 & 1.4 & 1.8 & 1.8 & 2.0 & 2.4 & 3.2 \\ 
SFRS - Real Layouts & 1M & \textbf{2.8} & \textbf{4.6} & \textbf{6.2} & \textbf{8.2} & \textbf{8.8} & \textbf{9.2} & \textbf{12.8} & \textbf{15.4} & \textbf{18.4}\\
Patch-SFRS - No Augmentation & 1M & 2.2 & 3.6 & 5.0 & 6.2 & 7.0 & 7.0 & 8.2 & 9.4 & 9.8 \\
Patch-SFRS - Real Layouts & 1M & 2.8 & 5.8 & 7.8 & 8.6 & 9.6 & 10.6 & 15.6 & 17.4 & 18.4 \\
Patch-NetVLAD - No Augmentation & 1M & 2.2 & 3.2 & 4.6 & 5.6 & 6.6 & 7.2 & 9.4 & 11.0 & 12.2\\
Patch-NetVLAD - Real Layouts & 1M & 2.2 & 3.4 & 3.8 & 4.6 & 5.0 & 5.0 & 7.2 & 8.0 & 8.4\\
\hline
SFRS - No Augmentation & Full &  0.8 & 2.0 & 2.4 & 2.6 & 3.2 & 3.8 & 5.0 & 6.2 & 7.6\\
SFRS - Gray Layouts & Full & 0.0 & 0.0 & 0.0 & 0.0 & 0.2 & 0.2 & 0.8 & 1.0 & 2.0\\ 
SFRS - Real Layouts & Full & \textbf{1.4} & \textbf{2.8} & \textbf{3.8} & \textbf{4.0} & \textbf{4.0} & \textbf{4.2} & \textbf{5.8} & \textbf{7.6} & \textbf{8.0}\\
Patch-SFRS - No Augmentation & Full & 3.4 & 3.6 & 4.2 & 4.8 & 5.0 & 5.6 & 6.4 & 7.6 & 7.6\\
Patch-SFRS - Real Layouts & Full & 2.2 & 3.0 & 3.6 & 4.4 & 4.4 & 4.6 & 6.8 & 7.6 & 8.0\\
Patch-NetVLAD - No Augmentation & Full & 2.4 & 3.4 & 4.2 & 5.4 & 6.0 & 6.2 & 7.6 & 8.2 & 8.4\\
Patch-NetVLAD - Real Layouts & Full & 1.8 & 3.2 & 3.8 & 4.0 & 4.2  & 4.4 & 5.2 & 5.4 & 5.4\\
\hline
\end{tabular}}
\caption{Results of Indoor-Ams validation set, evaluated on the full Outdoor-Ams and its subsets of 12k, 100k, and 1M images, corresponding to Fig 3(c)}
\label{table:4c}
\end{table*}

\begin{table*}[h!]
\centering
\resizebox{\textwidth}{!}{\begin{tabular}{ |c|c|c|c|c|c|c|c|c|c|c| }
\hline
\textbf{Model} & \textbf{Size} & \textbf{R@1} & \textbf{R@5} & \textbf{R@10} & \textbf{R@15} & \textbf{R@20} & \textbf{R@25} & \textbf{R@50} & \textbf{R@75} & \textbf{R@100}\\
\hline
SFRS - No Augmentation & 12k &  3.4 & 9.0 & 11.8 & 17.2 & 21.6 & 24.2 & 35.0 & 41.8 & 46.0\\ 
SFRS - Gray Layouts & 12k & 3.0 & 7.4 & 10.8 & 16.8 & 19.6 & 22.0 & 34.0 & 44.4 & 50.2 \\ 
SFRS - Real Layouts & 12k & \textbf{20.6} & \textbf{41.6} & \textbf{53.8} & \textbf{62.6} & \textbf{69.6} & \textbf{73.6} & \textbf{86.6} & \textbf{91.2} & \textbf{93.4}\\
Patch-SFRS - No Augmentation & 12k & 5.4 & 12.0 & 16.4 & 21.0 & 24.0 & 26.4 & 34.0 & 42.0 & 46.0\\
Patch-SFRS - Real Layouts & 12k & 10.8 & 24.6 & 34.2 & 43.8 & 48.20 & 54.0 & 72.6 & 84.4 & 93.4\\
Patch-NetVLAD - No Augmentation & 12k & 6.2 & 14.2 & 19.8 & 23.8 & 26.6 & 28.8 & 39.6 & 46.0 & 52.0  \\ 
Patch-NetVLAD - Real Layouts & 12k & 7.6 & 15.4 & 21.2 & 23.8 & 28.4 & 31.8 & 42.8 & 49.6 & 54.8 \\
\hline
SFRS - No Augmentation & 100k &  1.8 & 4.4 & 6.0 & 7.6 & 7.8 & 9.0 & 14.8 & 19.2 & 23.2\\ 
SFRS - Gray Layouts & 100k & 0.8 & 2.4 & 3.4 & 4.6 & 5.4 & 6.4 & 10.0 & 13.4 & 16.2 \\ 
SFRS - Real Layouts & 100k & 7.0 & \textbf{18.4} & \textbf{24.2} & 28.6 & \textbf{33.2} & 36.0 & 47.4 & 54.8 & 62.4\\
Patch-SFRS - No Augmentation & 100k & 4.6 & 7.6 & 10.4 & 11.4 & 13.2 & 14.2 & 18.8 & 22.0 & 23.2\\
Patch-SFRS - Real Layouts & 100k & \textbf{8.4} & 18.0 & 23.0 & \textbf{28.8} & 32.4 & \textbf{36.4} & \textbf{49.8} & \textbf{56.6} & \textbf{62.4}\\
Patch-NetVLAD - No Augmentation & 100k & 5.2 & 8.4 & 10.8 & 13.0 & 15.0 & 16.6 & 20.4 & 23.2 & 24.8 \\
Patch-NetVLAD - Real Layouts & 100k & 3.4 & 5.8 & 8.2 & 9.0 & 10.2 & 11.8 & 15.4 & 18.0 & 19.8 \\
\hline
SFRS - No Augmentation & 1M &  1.0 & 2.4 & 3.0 & 3.8 & 4.8 & 5.2 & 6.8 & 8.0 & 9.4 \\ 
SFRS - Gray Layouts & 1M & 0.2 & 0.6 & 1.8 & 2.2 & 2.6 & 2.6 & 3.2 & 4.4 & 5.2 \\ 
SFRS - Real Layouts & 1M & 2.8 & 6.6 & 8.2 & 9.8 & 11.8 & 13.6 & 19.2 & 22.8 & 25.4\\
Patch-SFRS - No Augmentation & 1M & 2.8 & 4.4 & 5.0 & 5.6 & 5.8 & 6.2 & 7.0 & 8.6 & 9.4 \\
Patch-SFRS - Real Layouts & 1M & \textbf{5.6} & \textbf{8.8} & \textbf{11.4} & \textbf{12.6} & \textbf{13.6} & \textbf{15.6} & \textbf{22.0} & \textbf{24.4} & \textbf{25.4} \\
Patch-NetVLAD - No Augmentation & 1M & 3.2 & 5.0 & 5.6 & 6.2 & 7.2 & 8.0 & 10.6 & 11.8 & 12.4  \\ 
Patch-NetVLAD - Real Layouts & 1M & 1.0 & 2.8 & 3.8 & 4.4 & 4.8 & 6.0 & 8.4 & 9.2 & 9.6 \\
\hline
SFRS - No Augmentation & Full & 1.2 & 2.2 & 3.6 & 4.0 & 4.6 & 4.8 & 5.8 & 8.0 & 9.8 \\ 
SFRS - Gray Layouts & Full & 0.0 & 0.0 & 0.0 & 0.2 & 0.2 & 0.4 & 0.4 & 0.8 & 1.4 \\ 
SFRS - Real Layouts & Full & 1.4 & 3.0 & 4.2 & 4.8 & 5.4 & 5.6 & 7.0 & 8.6 & 10.4 \\
Patch-SFRS - No Augmentation & Full & 3.2 & 4.4 & 5.4 & 5.6 & 5.8 & 6.2 & 8.0 & 8.6 & 9.8 \\
Patch-SFRS - Real Layouts & Full & \textbf{4.2} & \textbf{5.2} & \textbf{6.4} & \textbf{6.8} & \textbf{7.0} & \textbf{7.4} & \textbf{9.8} & \textbf{10.2} & \textbf{10.4}\\
Patch-NetVLAD - No Augmentation & Full & 2.4 & 3.4 & 4.0 & 4.0 & 4.6 & 4.8 & 6.6 & 8.8 & 9.4\\
Patch-NetVLAD - Real Layouts & Full & 1.4 & 1.8 & 2.6 & 3.2 & 3.2 & 3.2 & 4.4 & 4.8 & 5.0\\
\hline
\end{tabular}}
\caption{Results of Indoor-Ams test set, evaluated on the full Outdoor-Ams and its subsets of 12k, 100k, and 1M images, corresponding to Fig 3(d)}
\label{table:4d}
\end{table*}
\end{document}